\documentclass[runningheads]{llncs}
\usepackage[T1]{fontenc}
\usepackage{graphicx}
\usepackage{orcidlink}
\PassOptionsToPackage{hyphens}{url}
\usepackage{hyperref}

\begin{document}
    \title{From the Laboratory to Real-World Application: Evaluating Zero-Shot Scene Interpretation on Edge Devices for Mobile Robotics
    	\thanks{This preprint has not undergone peer review or any post-submission improvements or corrections. 
    		The Version of Record of this contribution is published in:
            Artificial Intelligence XLII. Lecture Notes in Computer Science.
            Springer Cham, 2026. DOI: \url{https://doi.org/10.1007/978-3-032-11442-6_21}}}

    \titlerunning{Scene Interpretation on Edge Devices}

    \author{Nicolas Schuler\inst{1,2}\orcidlink{0009-0007-4098-1244} \and
    Lea Dewald\inst{1}\orcidlink{0009-0004-8825-0545} \and
    Nick Baldig\inst{1,2}\orcidlink{0009-0002-8422-3406} \and
    Jürgen Graf\inst{1}\orcidlink{0000-0002-1354-0888}}

    \authorrunning{N. Schuler et al.}

    \institute{Laboratories for Cognitive Systems and Robotics -- Department of Computer Science -- Trier University of Applied Sciences, Trier, Germany\\
    \email{\{schulern,lxdw0338,n.baldig,j.graf\}@hochschule-trier.de} \and
    Department of Engineering -- Faculty of Science, Technology and Medicine -- University of Luxembourg, Kirchberg, Luxembourg}

    \maketitle

    \begin{abstract}
        Video Understanding, Scene Interpretation and Commonsense Reasoning are highly
        challenging tasks enabling the interpretation of visual information, allowing
        agents to perceive, interact with and make rational decisions in its
        environment.
        Large Language Models (LLMs) and Visual Language Models (VLMs) have shown
        remarkable advancements in these areas in recent years, enabling
        domain-specific applications as well as zero-shot open vocabulary tasks,
        combining multiple domains.
        However, the required computational complexity poses challenges for their
        application on edge devices and in the context of Mobile Robotics, especially
        considering the trade-off between accuracy and inference time.
        In this paper, we investigate the capabilities of state-of-the-art VLMs for the
        task of Scene Interpretation and Action Recognition, with special regard to
        small VLMs capable of being deployed to edge devices in the context of Mobile
        Robotics.
        The proposed pipeline is evaluated on a diverse dataset consisting of various
        real-world cityscape, on-campus and indoor scenarios.
        The experimental evaluation discusses the potential of these small models on
        edge devices, with particular emphasis on challenges, weaknesses, inherent
        model biases and the application of the gained information.
        Supplementary material is provided via the following repository: \hspace{1px}
        \url{https://datahub.rz.rptu.de/hstr-csrl-public/publications/scene-interpretation-on-edge-devices/}

    \end{abstract}

    \section{Introduction}\label{sec:introduction}

    The development of mobile cognitive systems offers great potential to create
    autonomous robotic platforms that can operate in dynamic and unstructured
    environments by using image sequences and context to enable decision-making.
    Visual commonsense and reasoning plays a crucial role in this, as it enables
    the systems to interpret visual information and make optimal decisions w.r.t.\
    utility based on it.
    By incorporating Visual Commonsense\cite{Zellers.2019}, a platform can reason
    over the relationships between objects and understand human behavior as the sum
    of their actions, working towards pareto optimality between safety, efficiency
    and comfort for the end user.
    This extends the capability of mobile platforms from just perceiving their
    environment to making sense of it in a more human-like way.
    For example, a service robot could not only identify a cup on a table but also
    understand that the person nearby is done taking his meal, starting an
    autonomous cleanup procedure.
    The potential of Visual Commonsense in mobile platforms is thus extensive.
    Use cases include autonomous vehicles, which in turn need to
    understand complex visual cues in urban environments, and service robots that
    assist in homes or hospitals, where subtle understanding of human needs is
    beneficial in terms of maximizing efficiency and comfort in parallel, enhancing
    human-robot interaction, cooperation and collaboration.

    Different approaches tackle these problems either by incorporating LLMs via
    external server and extensive post-processing with the mobile platform only
    serving as a sensory system\cite{ChenAnnieS.etal..2024}, or by utilizing
    domain-specific models, e.g.\ Convolutional Neural Networks (CNNs) or vision
    transformer on the edge device itself, with the application of commonsense
    reasoning VLMs on edge devices remaining limited\cite{Feroze2024,Farhan2025}.
    However, the integration of local solutions on edge devices are of particular
    interest for Mobile Robotics, were the accessibility of external services
    can not be guaranteed and might not be desirable. e.g.\ concerning privacy,
    and the zero-shot capabilities allow for open-domain usage, without being
    limited set of actions that will not adequately describe real world
    scenarios.
    In this paper, we aim to analyze the capabilities of such models for edge
    devices to describe a scene and the actors within on real-world data, with a
    focus on human action recognition, using a pre-trained VLM for zero-shot scene
    interpretation.
    In particular, we want to elucidate the challenges in bringing established
    model architectures to real-world applications in Mobile Robotics by
    incorporating the VLM in our architecture already used in the context of
    Mobile Robotics.

    The remainder of the paper is structured as follows:
    Section~\ref{sec:foundations-and-related-work} describes the foundation of
    Visual Commonsense and Video Understanding models, especially in regard to
    recent advancements with LLMs and VLMs.
    In addition, related work on applying such models to edge devices and Mobile
    Robotics are discussed, with an emphasis on centralized and de-centralized
    solutions.
    The proposed architecture is presented in
    Section~\ref{sec:integrated-zero-shot-video-understanding}, with the approach
    being experimentally evaluated in Section~\ref{sec:experimental-evaluation}.
    Finally, Section~\ref{sec:conclusion-and-future-work} concludes the work and
    discusses potential future applications and improvements.
    Supplementary material is provided via the following repository:
    \url{https://datahub.rz.rptu.de/hstr-csrl-public/publications/scene-interpretation-on-edge-devices/}

    \section{Foundations and Related Work}\label{sec:foundations-and-related-work}

    The foundations include different approaches to Visual Commonsense, especially
    utilizing LLMs and VLMs, which form the basis of the presented approach,
    common tasks, their respective datasets and metrics,
    see Section~\ref{subsec:different-approaches-to-visual-commonsense}.
    Section~\ref{subsec:vision-language-models} discusses various LLMs and VLMs,
    with an emphasis on small models for edge devices and Scene Interpretation.
    Following, the application of such models in the context of Mobile Robotics
    in the literature is discussed in
    Section~\ref{subsec:application-in-robotics-and-navigation}.

    \subsection{Different approaches to Visual Commonsense}\label{subsec:different-approaches-to-visual-commonsense}

    While humans are able to reason about observed entities and relations in
    between them, reasoning about actions and context from a single glance at a
    scene, this remains challenging for perceptive systems to this
    day\cite{Zellers.2019,Zhou.2024,ChenAnnieS.etal..2024}.
    This kind of understanding is essential for making rational decisions and
    provides an important basis for further development in Mobile Robotics.
    It exceeds mere object recognition and requires the ability to draw more
    complex conclusions from visual information that goes beyond the
    obvious\cite{Zellers.2019}.

    With the release of the Visual Commonsense Reasoning (VCR) task
    by~\cite{Zellers.2019} in 2019, this challenge was brought back into the
    spotlight.
    VCR aims to enable a human like understanding of visual scenes.
    The task contains to answer a multiple-choice question about an image
    (Q$\longrightarrow$A) and then to justify the given answer by answering another
    multiple-choice question  (QA$\longrightarrow$R), including the VCR
    dataset\cite{Zellers.2019}, containing 290,000 multiple-choice questions
    related to 110,000 film scenes\cite{Zellers.2019}.

    A recent approach to VCR, ViCor\cite{Zhou.2024}, aims to combine the strengths
    of Large Language Models (LLMs) and Vision Language Models (VLMs) and to
    compensate for their respective weaknesses.
    Visual Commonsense Reasoning problems were divided into two categories for this
    purpose: Visual Commonsense Understanding (VCU) and Visual Commonsense
    Inference (VCI).
    While VLMs are particularly good at reasoning and classifying visual
    content, LLMs are characterized by the fact that they can draw deeper
    conclusions that are not immediately apparent from the image content, such as
    cause-and-effect relationships and intentions of a person.
    Problems of this type fall into the category of VCI\@.
    ViCor uses pre-trained LLMs to determine the problem category.
    In addition, the LLMs act as controllers for the VLMs.
    As part of the VCU task, the model must recognize whether a text describes a
    concept or attribute from the given image.
    In contrast, solving VCI problems involves deriving new insights or
    explanations from the visual content, such as the purpose of an
    object.\cite{Zhou.2024}

    \subsection{Vision Language Models and Scene Representation}\label{subsec:vision-language-models}

    With the wide-spread success of LLMs, various closed- and open-source LLM
    models have been available to the community, e.g.\
    LLama3\cite{Grattafiori2024}, DeepSeek\cite{DeepSeekAI2024},
    Qwen\cite{Bai2023} and GPT4\cite{OpenAIetal..2024}.
    While these models report impressive results in various tasks, these
    results come at the cost of large-scale architectures (billions up to
    trillions trainable parameters), limiting the application of such models mostly
    to cloud-solutions.
    For the following section and the remainder of this paper, we will call models
    with up to a few dozen of billion parameters medium-sized and models with up to
    four billion models small, i.e.\ small language models (sLMs) or small vision
    language models (sVLMs).

    For medium-sized models,~\cite{Bai2023} provided a set of large-scale
    vision-language models (LVLMs) that can perceive and inference both text and
    images.
    The LLM Qwen-7B\cite{Bai2023} generates the desired responses
    according to the given prompts and performs various vision-language tasks,
    such as image labeling, question answering, text-based question answering and
    visual grounding\cite{Bai2023}.
    While a model size of seven billion parameters enables the usage of such models
    on workstations, edge devices place further computational restrictions on
    models, especially when other algorithms are to be run on the same device.

    Examples of such small models for image interpretation on edge devices include
    Moondream2\cite{vik2024}, and SmolVLM2\cite{Marafioti2025} consisting of
    1.93 and 2.2 billion parameters respectively, allowing these models to be
    efficiently run on edge devices.
    However, due to its size, such models are more likely to generate wrong captions
    and introduce biases in their answers\cite{Patnaik2025}.
    For a comprehensive survey on various sVLMs, see~\cite{Patnaik2025}.
    Regarding sLMs, as part of the wider LLama model family\cite{Grattafiori2024},
    Llama-3.2-3B and Llama-3.2-1B offer text-to-text sLMs for edge devices.
    For detailed recent survey on VLMs on edge devices, refer
    to~\cite{Sharshar2025}.

    Besides textual depictions of the scene, graph-based representation plays a
    crucial role in semantic scene representation.
    Scene graphs\cite{Chang.2022} represent a scene by temporal and spatial nodes
    and edges, which makes them well interpretable but requires extensive prior
    knowledge about the scene.
    Action Graphs\cite{HendriaWillyFitraetal..2023} also integrate the
    relationships between objects within and between frames as edge properties,
    where objects are represented as nodes and their interactions as edges.

    \subsection{Applications in Mobile Robotics}\label{subsec:application-in-robotics-and-navigation}

    For autonomous coordination in unstructured environments, systems must not only
    recognize objects, but also understand context-based intentions and make
    decisions proactively.
    Various approaches have been developed in recent years to tackle this
    challenge\cite{ChenAnnieS.etal..2024,ChenguangHuangetal..2023,Farhan2025,Zellers.2019,Zhou.2024}.

    The work of~\cite{ChenAnnieS.etal..2024} uses VLMs in walking robots to give
    them the ability to navigate autonomously through complex environments and
    overcome various obstacles - for example, as part of search operations in
    collapsed buildings.
    The robot must be able to climb over rubble, crawl through narrow gaps and find
    its way out of dead ends by retracing its steps.
    The robot should act appropriately in such circumstances, especially in
    unexpected situations.
    To make this possible,~\cite{ChenAnnieS.etal..2024} present the concept of
    VLM-Predictive Control (VLM-PC\cite{ChenAnnieS.etal..2024}).
    VLM-PC consists of two central components: the context-based adaptation to
    previous robot interactions and the planning of multiple capabilities into the
    future, including possible re-adaptation of plans.
    VLMs enable robots to independently perceive their environment, navigate and
    act in complex scenarios based on past interactions, future action strategies
    and extensive semantic knowledge - without having to rely on specialized
    technologies or external support, i.e.\@ assistance.
    The robot uses skills previously learned through reinforcement learning, such
    as walking, climbing or crawling, to efficiently navigate its
    environment\cite{ChenAnnieS.etal..2024}.

    VLMaps\cite{ChenguangHuangetal..2023} is a spatial map representation that
    directly combines pre-trained visual language features with a 3D reconstruction
    of the physical world.
    VLMaps can be created autonomously from robotic video recordings using standard
    exploration approaches and enable natural language indexing of the map, e.g.
    for the localization of landmarks or spatial references in relation to
    landmarks, without additional data acquisition or model fine-tuning.
    In particular, in combination with LLMs, VLMaps can be used to translate
    natural language commands into a sequence of open-vocabulary navigation targets
    that are directly localized in the map and can be shared by multiple robots
    with different equipment to create new obstacle maps on-the-fly, using a list
    of obstacle categories:
    For autonomous vehicles,~\cite{Farhan2025} utilize a combination of CNN-based
    incident detections, small local VLMs for textual scene descriptions and large
    cloud-based LLM to produce incident reports.

    \section{Integrated Zero-Shot Video Interpretation}\label{sec:integrated-zero-shot-video-understanding}

    While multimodal VLMs have shown remarkable results in various zero-shot vision
    tasks related to reasoning and understanding\cite{Li2025}, the performance of
    these architectures rely on being pre-trained on large-scale datasets in
    combination with a large number of trainable parameters.
    On the other hand, CNN- and efficient transformer-based
    architectures\cite{Luo2025} used for their low computational complexity on edge
    devices lack the strong generalization and zero-shot capabilities of the
    aforementioned VLMs.
    However, recent advancements allow for the utilization of local VLMs on
    edge devices\cite{vik2024,Marafioti2025}.
    To that end, this work presents an architecture leveraging a local VLM to allow
    their usage on edge-devices and in the context of Mobile Robotics.

    \subsection{Proposed Architecture}\label{subsec:proposed-architecture}

    The presented approach is split into two distinct systems: the mobile cognitive
    system with the edge device and a cloud-based solution for foundation model
    support.
    While the edge device utilizes small models for the initial scene description,
    object detection and tracking process, the cognitive system may communicate with
    the external system to gain further insides into its environment.
    While the agent is connected to the external system, personalized data is only
    processed by the local models, where external models can be used to further
    analyze the information generated by the models on the edge device, allowing
    for a privacy-preserving environment.

    Models on the edge device include a VLM for Scene Interpretation and further
    models for object detection and tracking.
    Using the generated model outputs, external cloud-based models too large to be
    run on local setups can further be queried to solidify the initial results.
    The described architecture is given in Fig~\ref{fig:arch_fig1}.
    An overview of the cognitive systems is given in the Appendix.
    For the purpose of the present work, the focus lies on the models used on the
    edge device of the cognitive system, especially the VLM.

    \begin{figure}
        \includegraphics[width=\textwidth]{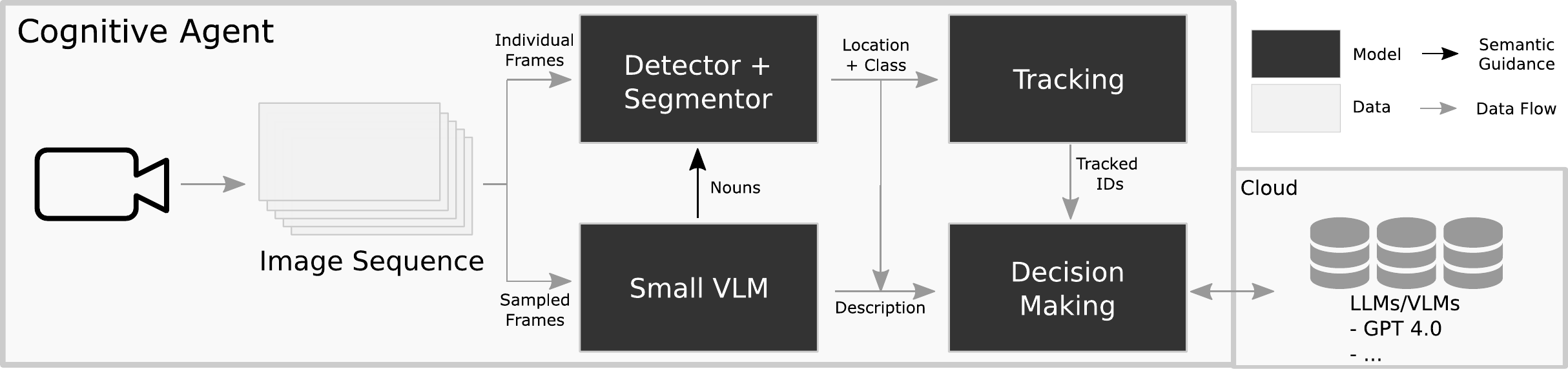}
        \caption{
            Architecture for Scene Interpretation on edge devices for Mobile Robotics.}
        \label{fig:arch_fig1}
    \end{figure}

    \subsection{Mobile Scene Interpretation and Guided Segmentation}\label{subsec:mobile-scene-understanding-and-guided-segmentation}

    The proposed pipeline works as follows: First, the local VLM  generates
    textual descriptions of the scene, given a sequence of images over the most
    recent time interval.
    Thus, the raw image data is only processed locally, preserving the privacy of
    the individuals when consulting larger inference models that might need to be
    run on an external server.

\begin{figure}
        \includegraphics[width=\textwidth]{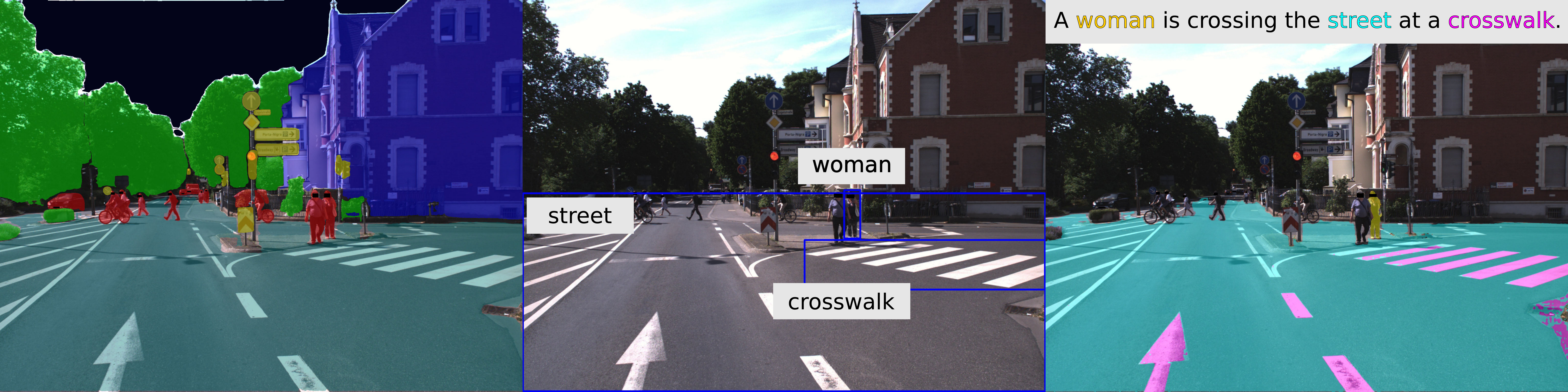}
        \caption{
            Example Result of the Proposed Pipeline.
            Left: The default segmentation given a fixed set of classes.
            Middle: The results of the first step of the pipeline with the generated
            VLM description, using the extracted nouns of the description for
            further object localization.
            Right: The final result of the Scene Interpretation and semantically-guided
            image segmentation.
            Note that the anonymization is only done for the purpose of this
            publication, the pipeline works with the raw data instead.}
        \label{fig:arch_fig2}
    \end{figure}

    The resulting description then can be used in various downstream tasks.
    For one, it might be used for further inferences, incorporating local or
    cloud-based LLMs.
    Within the local pipeline, the generated description can be used for
    semantically guided segmentation and tracking.
    To that end, the generated description is decomposed into nouns, which are in
    turn used for prompted zero-shot segmentation, giving additional insight into
    the generated scene description.
    An example of this process is given in Fig. \ref{fig:arch_fig2}, comparing the
    default segmentation for a given set of classes (e.g.\ cars, buildings,
    persons, bicycles, trees, traffic signs, street and sky) to the
    semantically-guided segmentation, focusing on elements important to the
    description only.
    The generated description `A woman is crossing the street at a crosswalk'
    yields the three nouns `woman', `street' and `crosswalk'.
    Note that the object detection and segmentation is done using zero-shot models,
    that is the used text prompts are not limited to arbitrary, pre-trained
    classes.
    In this particular setup, we utilize Grounded DINO\cite{Liu2023a} in
    combination with SAM\cite{Kirillov2023} for the task of zero-shot object
    detection and segmentation.

    \section{Experimental Evaluation}\label{sec:experimental-evaluation}

    The experimental evaluation analyzes the utilized VLM of the pipeline described
    in Section~\ref{sec:integrated-zero-shot-video-understanding} on a diverse
    dataset featuring various real-world outdoor and indoor scenarios.
    Due to the challenging task of open vocabulary video interpretation
    (see~\cite{Zhang2019,Hanna2021,Sun2022,Cheng2024,Kassab2025} for various open
    vocabulary evaluation metrics and their respective problems), we evaluate
    the quality of the results in terms of the concurrence of the generated scene
    description with a manually annotated description, see
    Section~\ref{subsec:experimental-setup} for more details.

    \subsection{Experimental Setup}\label{subsec:experimental-setup}

    The proposed pipeline is implemented in Python 3.12 using
    PyTorch\cite{Paszke2019} and Transformers\cite{Wolf2019}.
    As the VLM, we use the state-of-the-art model SmolVLM2\cite{Marafioti2025},
    which allows for real-time Scene Interpretation on edge devices using input
    video streams.
    The used dataset consists of 234 minutes of video data gathered in the german
    city of Trier by our laboratory.
    We differentiate three distinct domains: \textit{Campus Indoor},
    \textit{Campus Outdoor} and \textit{City}, with a split of 107 minutes, 74
    minutes and 53 minutes respectively.
    Examples from the dataset and the used cognitive systems for the recording are
    given in the Appendix.
    The videos are split into five second clips that are manually annotated and
    fed into a VLM in order to describe the main action contained in the clip.
    The annotater of the video clips is given example outputs of the VLM to match
    their annotation to the length and style of the VLM\@.

    The generated description were evaluated by human experts for correctness, that
    is the accordance of the generated description with a manual evaluation of the
    sequence.
    The evaluation process is as follows.
    The generated description for a given sequence is compared to its
    human-generated baseline description.
    To quantify the results, the generated descriptions are evaluated in three
    sub-categories: \textit{Action} (relations), \textit{Agent} and
    \textit{Object} (entities).
    If a VLM-generated description contains the correct action, the sub-category
    \textit{Action} is considered correct.
    \textit{Agent} is considered correct if the correct type and number of agents
    is contained in the description.
    Finally, \textit{Object} is considered correct, if the description contains the
    correct type and number of objects relevant to the depicted action.
    If all three sub-categories are correct, the entire description is considered
    \textit{Correct}, meaning the main action of the sequence is correctly
    described, including the involved agents and utilized objects.
    E.g.\ if a sequence displays the action
    `A man and a woman are crossing the street at a crosswalk.', the generated
    description `A woman is crossing the street at a crosswalk.' is evaluated as
    \textit{Action}: correct, \textit{Agent}: false and \textit{Object}: correct,
    leading to an overall evaluation of \textit{Correct}: false, since the action
    and relevant objects were identified correctly but not all agents are
    mentioned (multiple people `a man and a woman' versus one `woman').
    Note that by default we do not consider empirical characteristics of the person
    when evaluating the generated descriptions,
    i.e.\ `A woman is crossing the street at a crosswalk.' and
    `A man is crossing the street at a crosswalk.' are considered equivalent.
    We note and discuss the difficulties in this approach in
    Section~\ref{subsec:discussion}, since the inherent subjectivity in human
    evaluation poses a significant challenge.
    However, due to the lack of reliable metrics, see
    Section~\ref{subsec:discussion}, for this task in the context of practical
    applications, we feel justified in this decision.

    In addition to the human evaluation, we utilize two popular semantic similarity
    scores to evaluate the generated descriptions.
    \textit{BERTScore}\cite{Zhang2019} utilizes BERT\cite{Devlin.2019} to calculate token
    similarity between the input pairs, while STSB DistilRoBERTa\cite{UKPL2024}
    uses cross encoding and similarity calculation between the two provided
    descriptions for its \textit{Sentence Similarity} scoring.
    For the purpose of this paper, we use these to evaluate the usefulness of such
    metrics within our domain, comparing the results of automated metrics to the
    manual evaluation results.
    To that end, we calculate the correlation coefficient \textit{R} between the
    automated metric and the manual evaluation \textit{Correct} using
    point-biserial correlation coefficient\cite{Tate1954}.
    In addition, the percentage of matches between the human and manual evaluation,
    in the following called \textit{Match \%}, is calculated.
    The matches are calculated by categorizing the sequences using a threshold for
    the given metric.
    If the sequence is rated over the threshold, the generated description is
    considered \textit{Correct}.
    The used threshold value is calculated using optimization on the generated
    metric values.

    \subsection{Experimental Results}\label{subsec:experimental-results}

\begin{figure}
        \includegraphics[width=\textwidth]{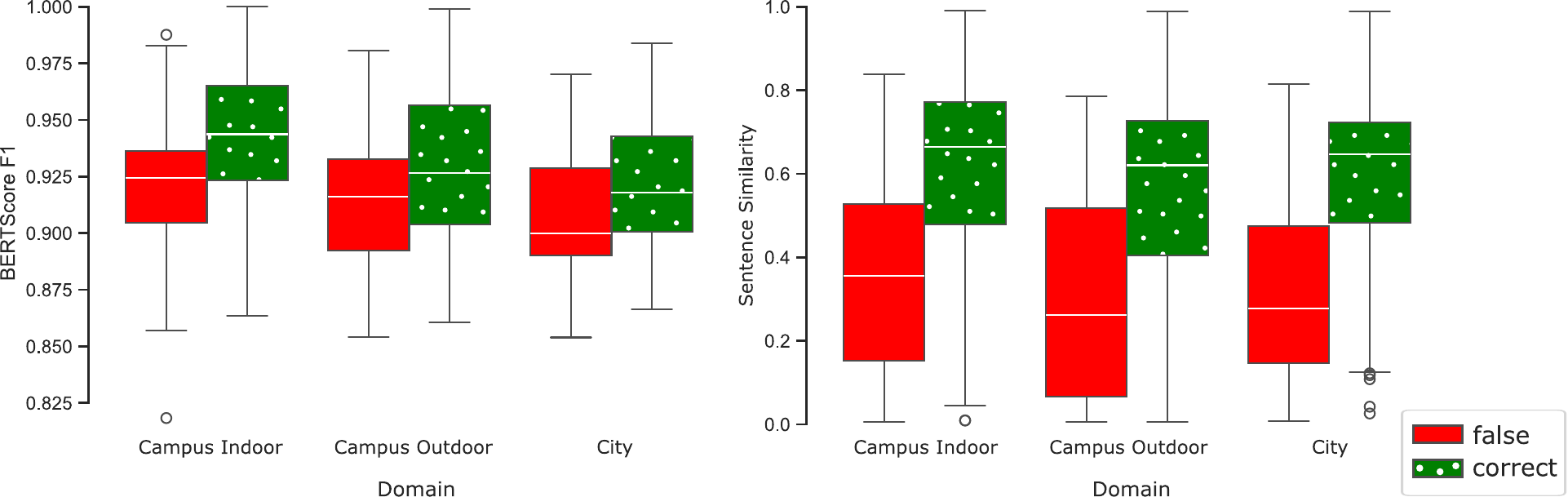}
        \caption{
            Boxplots for the two automated metrics \textit{BERTScore} and
            \textit{Sentence Similarity} by domain.
            Metrics are further grouped by the manually evaluated correctness of the
            generated descriptions compared to the ground truth.}
        \label{fig:exp1_fig2}
    \end{figure}

\begin{figure}
        \includegraphics[width=\textwidth]{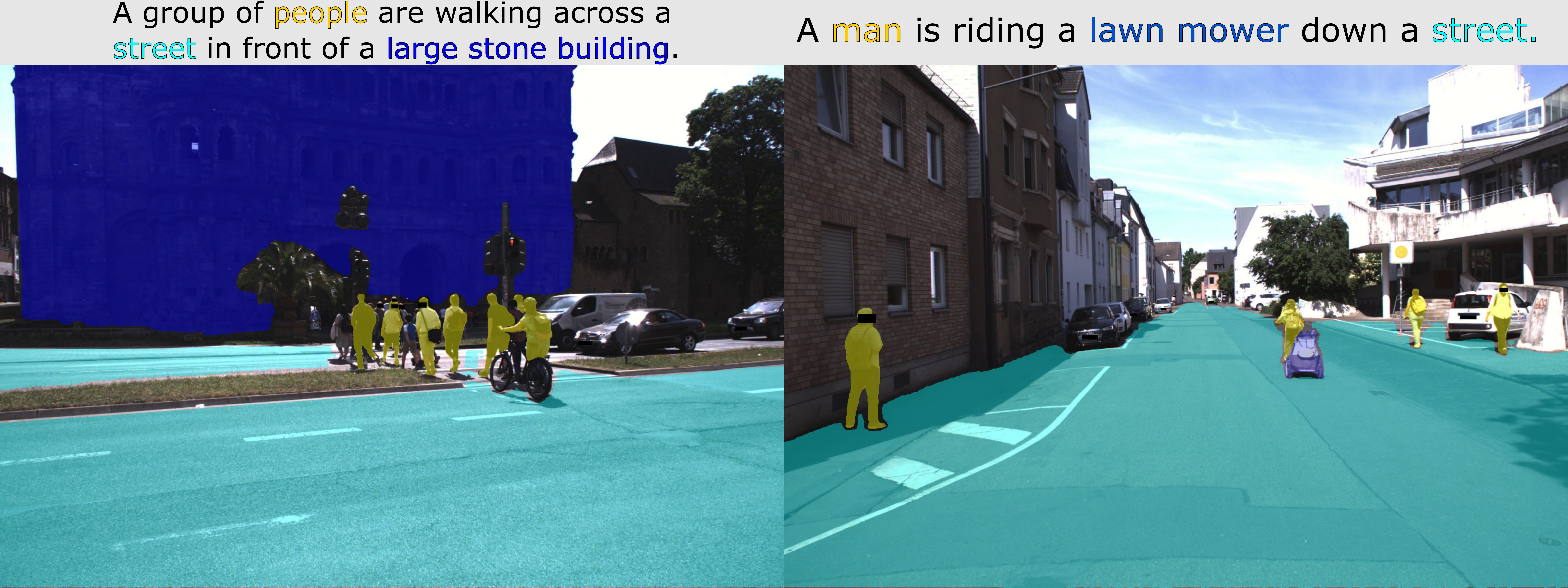}
        \caption{
            Two examples of a correct and a partially correct description.
            Using semantically-guided segmentation to highlight important elements of
            the generated description.
            Left: Correct description of the scene.
            Right: Partially correct description where the object associated to the
            action is considered incorrect (`lawn mower' instead of
            `bicycle with a child trailer').
            Note that the anonymization is only done for the purpose of this
            publication, the pipeline works with the raw data instead.}
        \label{fig:exp1_fig1}
    \end{figure}

    Results of the evaluation are given in Table~\ref{fig:exp1_tab1}, for a
    qualitative example the results see Fig.~\ref{fig:exp1_fig1}.
    The results are grouped by domain (\textit{Campus Indoor},
    \textit{Campus Outdoor} and \textit{City}).
    In total, 65.4\% of all generated descriptions are considered correct.
    Between the different domains, the domain \textit{Campus Indoor} has the lowest
    correctness score with 53.3\% and the domain \textit{City} the highest
    with 79.6\%.
    In the evaluated sub-categories \textit{Action}, \textit{Agent} and
    \textit{Object}, \textit{Agent} has the highest correctness with 93.7\%,
    compared to 78.9\% and 83.2\% for \textit{Action} and \textit{Object}
    respectively.
    Concerning the used automated metrics, i.e.\ \textit{BERTScore} and
    \textit{Sentence Similarity}, \textit{BERTScore} has a lower correlation
    coefficient for all domains versus \textit{Sentence Similarity}, with an R
    value of 0.229 to 0.483.
    This is further emphasized by the percentage of accordance between the
    manual evaluation and automated metrics, with 67.1\% for \textit{BERTScore}
    versus 73.6\% for \textit{Sentence Similarity}.
    Fig.~\ref{fig:exp1_fig2} gives the boxplots for \textit{BERTScore} and
    \textit{Sentence Similarity} split by domain and ground-truth, that is manually
    evaluated correctness.

    \begin{table}
        \caption{
            Results of the evaluation of SmolVLM2\cite{Marafioti2025} within the
            proposed pipeline.
            Including human evaluation (\textit{Correct},
            \textit{Action}, \textit{Agent} and \textit{Object}) and automated
            similarity metrics (\textit{BERTScore} and \textit{Sentence Similarity}).
            The generated description of the individual sequence is compared to the
            annotated ground truth.
            A description is considered correct if it contains the main action, the
            type and number of acting agents and the involved objects.
            For a detailed description of the used metrics see
            Section~\ref{subsec:experimental-setup}.}
        \label{fig:exp1_tab1}
        \includegraphics[width=\textwidth]{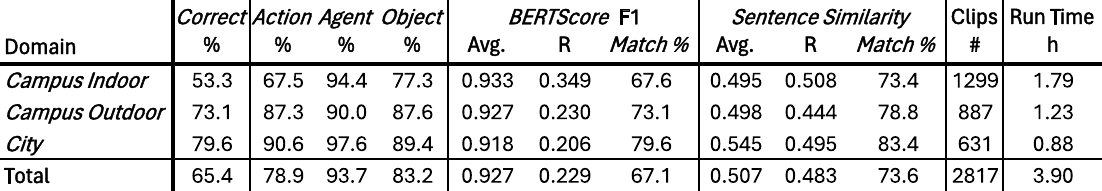}
    \end{table}

    \subsection{Discussion}\label{subsec:discussion}

    The following discussion focuses on three distinct aspects.
    First, the results of the experimental evaluation are discussed, with an
    emphasis on lessons learned when applying a state-of-the-art VLM to our setup
    within the context of Mobile Robotics.
    Following, the difficulties inherent to manual as well as automated evaluation
    are elucidated.

    \subsubsection{Applying Vision Language Models to Real-World Scenarios}\label{subsubsec:applying_vlm_to_real_world}

    The results highlight the capabilities of SmolVLM2 for usage within our
    pipeline in principle.
    However, the difference in correctness between domains is considerable.
    This difference might be explained by two factors.
    First, the domain with the lowest correctness is \textit{Campus Indoor}.
    This domain contains a more diverse and complex set of activities, e.g.\
    `A woman is putting dishes into a dishwasher.',
    `A man is giving an apple to another person.',
    `A woman is picking up a pair of scissors from a table.',
    `A person is writing on whiteboard.',
    `A person is climbing a ladder.'.
    On the other hand, while the domains \textit{Campus Outdoor} and \textit{City}
    can also contain complex actions, e.g.\
    `A group of people are taking measurements on a street.', a lot of the depicted
    actions are more simplistic in nature, e.g.\
    `A man is walking on the sidewalk.', `A car is driving down a road.'.
    For the second factor, refer to the next section and the discussion on the
    difficulties in evaluating the results concerning the different domains.

    We want to further discuss biases in the generated descriptions of the used
    model.
    For one, the model struggles to differentiating certain actions that are
    closely related but decidedly different, e.g.\ the model uses `sitting down'
    to describe both the action of `sitting down' and `standing up'.
    Another point is a heavy bias regarding certain types of objects present in a
    scene.
    E.g.\ if a whiteboard is present in the scene, the model will include the
    whiteboard into the action, even if not relevant to the action, leading to
    descriptions like
    `A person is standing on a ladder and is writing on a whiteboard.' instead of
    the ground truth of `A person is climbing a ladder.'.
    This means that care has to be taken when using the model within specific
    domains and carefully evaluating the potential relevant biases within that
    domain and filtering the generated output according to this prior knowledge can
    help to stabilize the predictions.

    Another important lesson is the utilization of the gained information.
    In our current setup, the generated scene description relates to the past
    five seconds.
    On top of that, depending on the complexity of the scene and system load,
    another one to three seconds are needed for the inference itself.
    Thus, the delay between an action happening and the cognitive agent
    receiving the information might be as high as eight seconds.
    While this delay can be reduced to a degree by changing the evaluated time
    interval, the basic problem remains and must be considered.
    Consequently, the cognitive systems should use the gained information either
    only in contexts that are not time-sensitive, e.g.\ within assistance tasks
    that are not time-critical or to document incidents within more time critical
    environments.
    For example, within our domain \textit{City}, the gained information can be
    used to generate dynamic, cooperative maps, highlighting points of interest,
    e.g.\ sections with a lot of pedestrian traffic, streets with a high number of
    cars suddenly driving on the road due to hidden driveways.

    \subsubsection{Inherent Difficulties Regarding Evaluation}\label{subsubsec:inherent_difficulties_regarding_evaluation}

    The challenges in manual evaluation are twofold.
    For one, the task of generating the ground truth manual annotations concerning
    the main action for a given scene can be ambiguous, especially depending on the
    domain.
    Within our domain \textit{Campus Indoor}, the main action is usually clearly
    defined.
    The actions we recorded within that domain are mostly work related:
    working in an office, carrying something, working at a computer, talking to
    someone, or are related to movement: walking, sitting, taking the
    stairs et cetera.
    While these actions can be described vastly differently by a human annotator,
    e.g.\ `A man is sitting at a computer.' versus `A man is working in an office.'
    describing the same sequence, the difference in these descriptions can still be
    easily identified and evaluated correctly when comparing it to the generated
    output of the VLM, e.g.\ `A person is sitting at a table with a computer.',
    which would be considered as a correct description.
    This is not true for the domain \textit{City}, and to a lesser extent the
    domain \textit{Campus Outdoor}:
    Here, due to the open nature of the domains, e.g.\ large groups of people,
    large spaces and different types of road users, it can be difficult to identify
    a main action of a sequence.
    For example, all following sentences describe the same scene:
    `A busy intersection with cars driving and people crossing the road.',
    `A black car is crossing a busy intersection.',
    `People are crossing the street at a crossroad in front of a large building.',
    `A bus is stopping at a traffic light with people crossing the road.'.
    For the purpose of this paper, we consider all four of these descriptions
    \textit{Correct} if there is no one critical action identified within the
    sequence.
    That is, if a bike is crossing the street right in front of the cognitive
    agent, this would be the critical action and all descriptions given
    above would be false.
    If the cognitive agent is simply waiting at an intersection, with various cars
    waiting in front of it, all descriptions above are considered correct.

    Regarding the automated metrics, various works discuss the problems inherent in
    such methods.
    \textit{BERTScore}\cite{Zhang2019} displays significant biases concerning
    social, age, race, gender, religion and physical
    appearances\cite{Hanna2021,Sun2022}.
    More recent works explore probabilistic methods to evaluate semi- and
    open-vocabulary tasks like commonsense reasoning\cite{Cheng2024} and
    scene representation\cite{Kassab2025}.
    While these probabilistic approaches show promising
    results\cite{Cheng2024,Kassab2025} over the previous similarity measures,
    they are currently either limited to a question-answer setup\cite{Cheng2024} or
    to a limited task such as scene representation\cite{Kassab2025} where the
    labels of the objects to be segmented can be reasonably expected to be
    exhaustive, encompassing a complete set of synonymous and similar objects.
    Thus, these methods are not yet applicable to our task and we utilize more
    traditional similarity metrics.
    As shown in Fig.~\ref{fig:exp1_fig2}, both metrics correlate with the manual
    evaluation.
    However, the overlap between quartiles, which is particularly pronounced for
    BERTScore, shows the difficulties in relying on these metrics only to evaluate
    the performance of the model.
    Additionally, the accordance with the manual evaluation depends on the domain
    in question and thus has to be checked for every new domain or when
    encountering new actions within a given domain.

    \section{Conclusion and Future Work}\label{sec:conclusion-and-future-work}

    The present paper discussed and highlighted the challenges in applying current
    state-of-the-art VLMs to the task of Scene Interpretation for Mobile Robotics
    on edge devices.
    In particular, we demonstrated the strong capabilities of these models when
    used in zero-shot tasks on unknown domains in real-world scenarios.
    Finally, we discussed the challenges in applying such models to the real
    world, highlighting the difficulties in evaluating manual and generated
    descriptions and inadequacies in using current automated metrics for this task.

    For future work we want to highlight the need for better metrics for
    open-vocabulary tasks and future research into reliable evaluation of real
    world domains for these tasks, especially to reduce the need for manual
    evaluation.

    \begin{credits}
        \subsubsection{\discintname}
        The authors have no competing interests to declare.
    \end{credits}

    \section*{Appendix}\label{sec:appendix}

\begin{figure}
        \includegraphics[width=\textwidth]{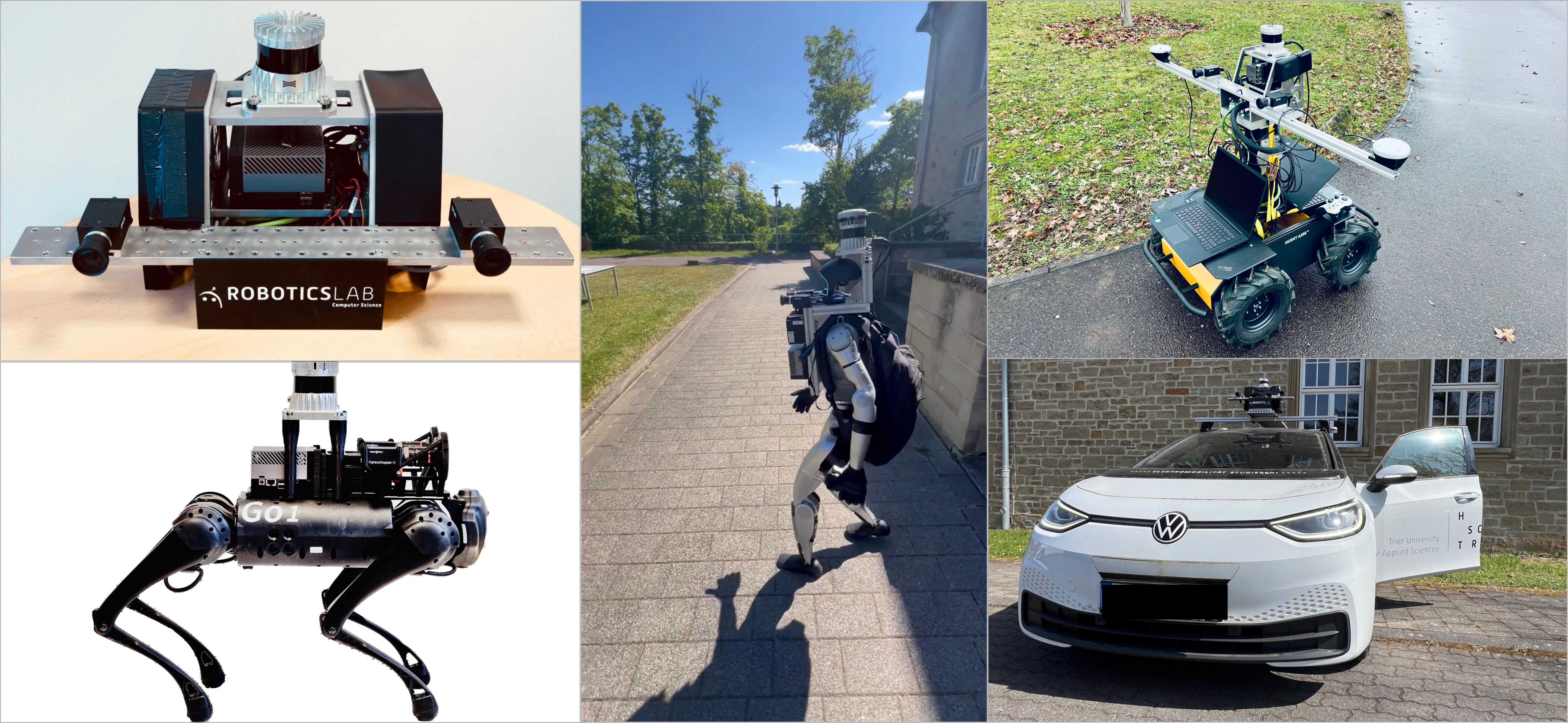}
        \caption{
            The different cognitive systems used within our laboratory, utilizing the
            same sensory system (top left) used for data acquisition in the present
            paper.
        }
        \label{fig:appendix_fig1}
    \end{figure}

    The used dataset was recorded using different cognitive systems, see
    Fig.~\ref{fig:appendix_fig1}.
    All platforms share the same sensor system that is only adapted to the given
    requirements w.r.t.\ the mechanical constraints.
    We utilize stereo cameras, LiDAR and a GNSS/INS module, with an
    NVIDIA Jetson AGX as the edge device.

\begin{figure}
        \includegraphics[width=\textwidth]{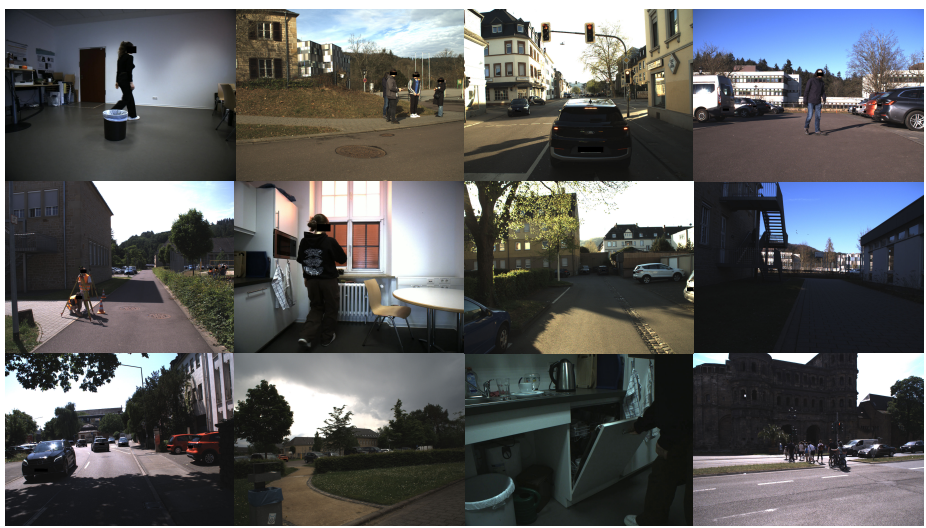}
        \caption{
            Various examples from the used dataset, captured by a variety of cognitive
            agents, including city and campus scenes, featuring
            difficult lighting conditions, crowded scenes and complex scenarios.
            Note that the anonymization is only done for the purpose of this
            publication, the pipeline works with the raw data instead.}
        \label{fig:appendix_fig3}
    \end{figure}

    The dataset used in this paper consists of a combination of various outdoor and
    indoor scenarios in real-world situations, see Fig.~\ref{fig:appendix_fig3} for
    a short overview.

\end{document}